%% file: main.tex
\documentclass[conference]{IEEEtran}
\IEEEoverridecommandlockouts
\usepackage{cite}
\usepackage{amsmath,amssymb,amsfonts}
\usepackage{algorithmic}
\usepackage{graphicx}
\usepackage{textcomp}
\usepackage{xcolor}

\usepackage{multirow}
\usepackage{bbm}
\usepackage{float}
\usepackage{gensymb}

\def\BibTeX{{\rm B\kern-.05em{\sc i\kern-.025em b}\kern-.08em
    T\kern-.1667em\lower.7ex\hbox{E}\kern-.125emX}}
\begin{document}

\title{CLRGaze: Contrastive Learning of Representations \\ for Eye Movement Signals}

\author{\IEEEauthorblockN{
Louise Gillian C. Bautista}
\IEEEauthorblockA{\textit{Department of Computer Science} \\
\textit{University of the Philippines}\\
Quezon City, Philippines \\
lcbautista1@up.edu.ph}
\and
\IEEEauthorblockN{
Prospero C. Naval, Jr.}
\IEEEauthorblockA{\textit{Department of Computer Science} \\
\textit{University of the Philippines}\\
Quezon City, Philippines \\
pcnaval@dcs.upd.edu.ph}}

\maketitle

\begin{abstract}
Eye movements are intricate and dynamic biosignals that contain a wealth of cognitive information about the subject. However, these are ambiguous signals and therefore require meticulous feature engineering to be used by machine learning algorithms. We instead propose to learn feature vectors of eye movements in a self-supervised manner. We adopt a contrastive learning approach and propose a set of data transformations that encourage a deep neural network to discern salient and granular gaze patterns. This paper presents a novel experiment utilizing six eye-tracking data sets despite different data specifications and experimental conditions. We assess the learned features on biometric tasks with only a linear classifier, achieving 84.6\% accuracy on a mixed dataset, and up to 97.3\% accuracy on a single dataset. Our work advances the state of machine learning for eye movements and provides insights into a general representation learning method not only for eye movements but also for similar biosignals.
\end{abstract}


\input{intro}

\input{method}

\input{data}

\input{experiment}
\input{results}

\input{conclusion}

\bibliographystyle{IEEEtran}
\bibliography{refs}

\end{document}

%% file: intro.tex
\section{Introduction}

Eye movements have long been used in Cognitive Science to understand how people think and perceive \cite{cognition-konig-2016, duchowski, etra, review-graphs-coutrot}.
For example, longer fixations can indicate information processing \cite{duchowski}, rapid saccades 
may imply a viewer's excitement \cite{saccade-arousal}, and microsaccades 
may indicate higher cognitive load \cite{etra}.
Beyond cognitive signals, researchers have also found idiosyncratic patterns in eye movements, spurring research into eye movement biometrics \cite{rigas-biometrics, emvic, lpitrack}.

Due to the breadth of information that can be extracted from gaze behavior, there is a wide array of studies on eye movements for use in domains such as education, safety, and healthcare \cite{duchowski, review-graphs-coutrot}.
However, eye movements can be difficult to process because of factors that affect gaze behavior including the stimuli used, tasks given, and eye-tracker specifications. Researchers then have to carefully select their methodologies to emphasize the patterns specific to their use-case. For example, formulas for extracting features have to be tuned \cite{eye-movement-features}, and areas-of-interests have to be manually defined \cite{scanpaths, review-graphs-coutrot}.
However, these methods may not generalize well to other data sets and use-cases because of their dependence on the stimuli used, prior beliefs, and data specifications (e.g. sample length, sampling frequency).

\begin{figure}[t]
    \centering
    \includegraphics[width=\columnwidth]{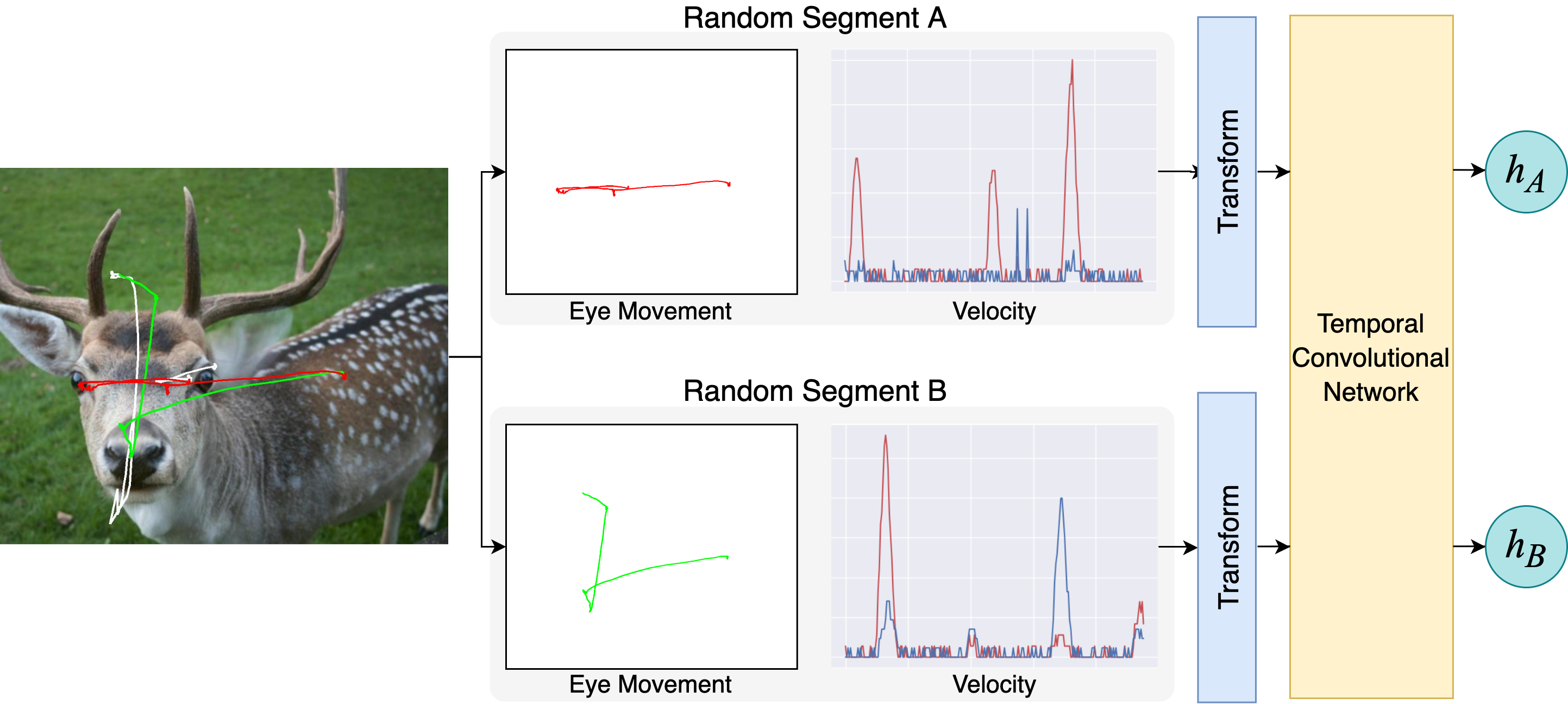}
    \caption{Overview of CLRGaze. For any eye movement sample, random segments are taken and treated as velocity signals. These undergo transformations and are fed into an encoder as part of a batch. The encoder is then trained to discriminate segments that originated from the same signal, in effect learning meaningful abstract representations of eye movements.}
    \label{fig:catch}
\end{figure}

We aim to do away with this arduous task of handcrafting features and manually selecting representation methods.
We instead automatically encode eye movements into feature vectors that accurately capture the salient and granular characteristics of gaze behavior. Any new eye movement sample can be mapped to this vector which can then be used for downstream tasks. The goal is to (1) encode high-resolution data such that even the minute movements are accounted for and (2) make it easier to apply machine learning algorithms to eye movement data for real-world applications.

We improve upon previous work \cite{gazemae} by taking a self-supervised contrastive learning approach, where a convolutional neural network (CNN) learns abstract representations of the data by being exposed to a large number of similar (positive) and dissimilar (negative) examples.
We follow SimCLR \cite{simclr}, a contrastive learning framework for images. In this framework, representations are learned by comparing and contrasting different views (i.e. transformations) of images.

We port this methodology to the signal domain, specifically applying it to eye movements (Fig. \ref{fig:catch}). We propose a set of signal cropping and transformations that encourage the network to discern important patterns in eye movements. To assure the efficacy of our proposed methodology, we conduct our experiment on six eye-tracking data sets. The joint data set consists of 45,755 eye movement trials from 143 subjects. After training the contrastive CNN on the eye movements, we evaluate the learned representations through inter and intra-dataset biometric tasks with only a linear classifier. Despite having different specifications, our method works well across these data sets, achieving 84.6\% accuracy across all samples and up to 97.3\% accuracy on a single data set. We also show that our model generalizes to unseen samples. 

Our contributions are as follows: (1) we apply contrastive representation learning to eye movement signals, (2) we propose a set of signal data transformations to aid contrastive learning, (3) we demonstrate the validity of our method by conducting experiments on six data sets, and (4) we achieve superior accuracies and establish new baselines on biometric tasks.




%% file: method.tex
\section{Methodology}

\subsection{Contrastive Representation Learning}

Our methodology is largely based on SimCLR \cite{simclr}.
For any image $x \in \mathbbm{R}^{d_x}$ in a mini-batch of size $N$, two different random image augmentations are performed (e.g. cropping, blur, color distortion) to obtain two new sub-images $x_a$ and $x_b$ that form a positive pair. This doubles the mini-batch size to $2N$, where each image now has one positive example and $2(N-1)$ negative examples.

A CNN $f$ encodes this mini-batch $\{x_k\} \in \mathbbm{R}^{d_x}$ to their representations $\{h_k\}$ in a learned latent space $\mathbbm{R}^{d_h}$. The representations are further mapped by a nonlinear projection head $g$ to a small feature vector $\{z_k\} \in \mathbbm{R}^{d_z}$ with which their intra-batch similarities are calculated. The encoder $f$ and projection head $g$ are jointly optimized such that the similarity between positive pairs are maximized while that of negative pairs are minimized. Specifically, they minimize the normalized temperature-scaled cross entropy loss (NT-Xent). For any data point $i$ and $j$ in the mini-batch, NT-Xent is computed as follows:

\begin{equation}
    \ell_{i,j} = -log \frac{exp(sim(z_i, z_j)/\tau)}{\sum_{k=1}^{2N} \mathbbm{1}_{[k \neq i]} exp(sim(z_i, z_k)/\tau)}
    \label{eq:nt-xent}
\end{equation}

where $sim$ is the cosine similarity between two projections $sim(z_i, z_j) = z_i^T z_j / \Vert z_i\Vert \Vert z_j \Vert$, $\mathbbm{1}$ is an indicator function evaluating to $1$ when the data point is not being compared to itself, and $\tau$ is a temperature parameter that scales the similarity values.
By training on larger batches and many iterations, the network is exposed to more positive and negative examples. Thus, allowing it to form richer abstractions about the data.

\subsection{CLRGaze}
\label{sec:clrgaze}
We take inspiration from SimCLR and apply this framework to eye movement signals. We port their methodology to the signal or 1D time-series domain by performing a set of data transformations analogous to augmentation techniques for images or 2-D data.


\begin{figure}[ht]
    \centering
    \includegraphics[width=\columnwidth]{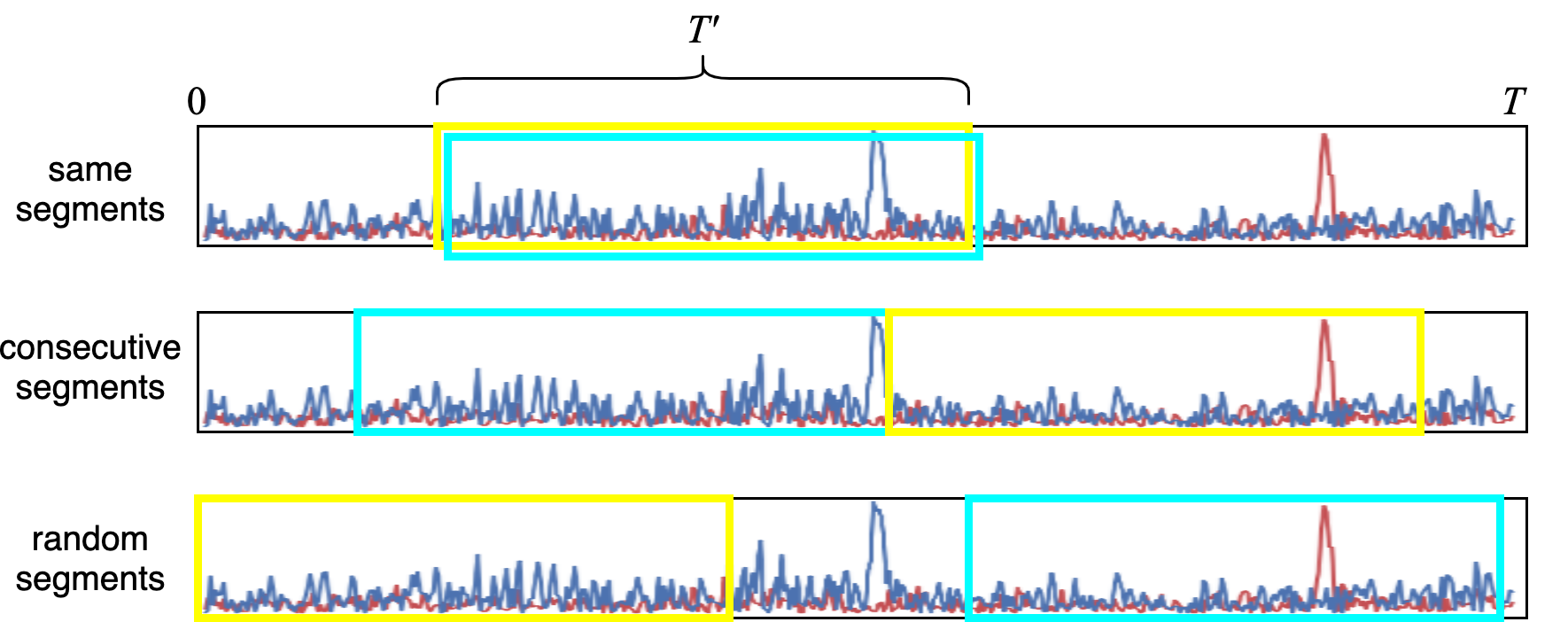}
    \caption{Cropping methods applied to each signal to form a positive pair.}
    \label{fig:cropping-method}
\end{figure}

\begin{table}[hb!]
    \centering
    \caption{The set of signal transformations used in CLRGaze. For each cropped segment, we randomly choose one with uniform probability (11.11\%) and apply to the signal.}
    \begin{tabular}{p{2.6cm}|p{5.1cm}}
         \textbf{Transformation} & \textbf{Description} \\
         \hline
         \hline
         1. Identity & No transformation. \\
         \hline
         2. Dropout & Randomly zero out 20\% of time points in the signal in both x and y dimensions. \\
         \hline
         3. Chunk Drop. & Zero out a 20\% chunk of the signal in both x and y dimensions. \\
         \hline
         4. Alternate Drop. & Alternately zero out time points in both x and y dimensions. \\
         \hline
         5. Channel Drop. & Zero out either the x or y dimension. \\
         \hline
         6. Gaussian Noise & Apply additive noise sampled from $\mathcal{N}(0, 0.5)$ \\
         \hline
         7. Drop. \& Noise & Randomly zero out 20\% of time points and apply additive noise sampled from $\mathcal{N}(0, 0.5)$. \\
         \hline
         8. Chunk Copy & Replace a 20\% chunk of the signal with a different 20\% chunk within the same signal. \\
         \hline
         9. Chunk Swap & Swap two disjoint 20\% chunks of the signal. \\
         \hline
         \multicolumn{2}{l}{Chunk transforms were partially inspired by \cite{chorowski}.}
    \end{tabular}
    \label{tab:data-transformations}
\end{table}

A sample signal $x \in \mathbbm{R}^{(2, T)}$ is a time-series with arbitrary length $T$ and 2 channels corresponding to the x and y plane Each time step is an estimated position of a viewer's gaze on a screen. To produce a positive pair $(x_a, x_b)$ from $x$, we must obtain two different views or segments $x_a, x_b \in \mathbbm{R}^{(2, T')}$ where $T' \leq T$ is a fixed input length. We do this by randomly selecting any of the three cropping methods visualized in Figure \ref{fig:cropping-method}.

Given $T'$ and a randomly selected time point $t$ where $t \leq T' \leq T$, the methods are as follows: In (1) \textit{Same}, the two segments are identical ($x_a = x_b = x_{t:t+T'}$). In (2) \textit{Consecutive}, the two segments are consecutive portions of the signal, i.e. $x_a$ immediately proceeds \textit{or} precedes $x_b$ (e.g. $x_a = x_{T'-t:t}$ and $x_b = x_{t:t+T'}$). In (3) \textit{Random}, the two segments come from any random portion of the signal (e.g. $x_a = x_{t_1: t_1+T'}$ and $x_b = x_{t_2: t_2+T'}$).
After cropping, we separately apply, with uniform probability, one out of nine transformations listed in Table \ref{tab:data-transformations}  to both segments. The transformations alter or destroy the signal encouraging the network to find unique patterns and be robust to noise.

\begin{figure*}[h]
    \centering
    \includegraphics[width=\textwidth]{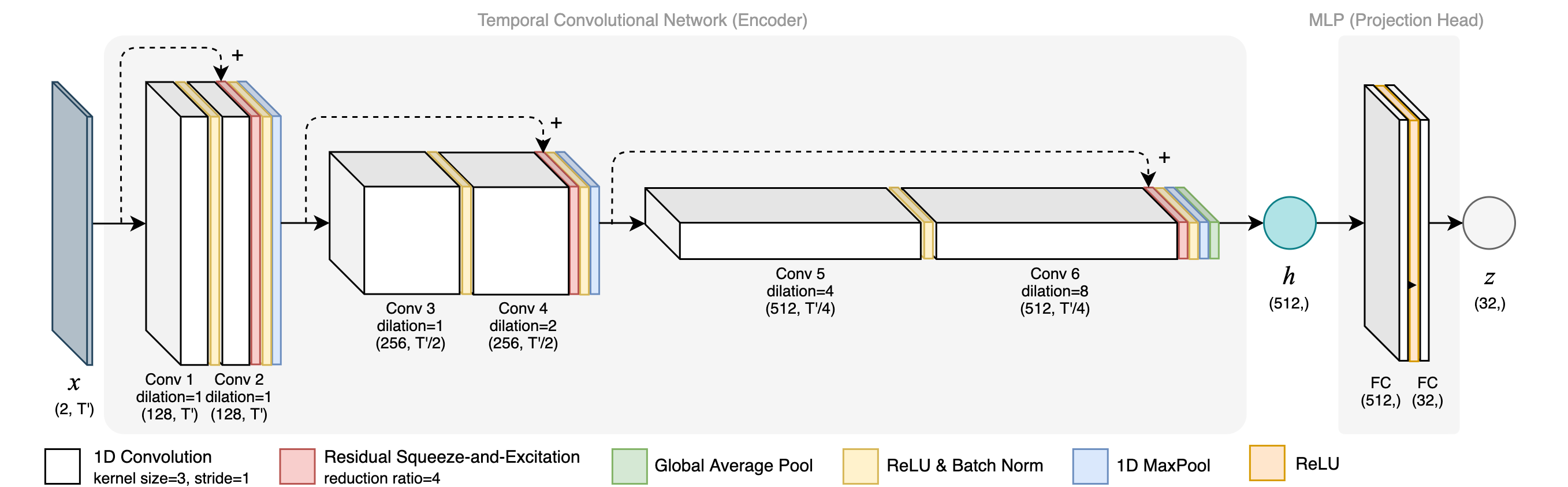}
    \caption{Overview of the TCN encoder and the MLP projection head used in CLRGaze, with 2,147,072 parameters. A mini-batch $\{x_k\}$ of transformed segments $x \in \mathbbm{R}^{(2, T')}$ passes through this network. NT-XEnt loss is calculated on $z$, while the final representation $h$ is taken as the output of the Global Average Pooling (GAP) layer.}
    \label{fig:encoder-architecture}
\end{figure*}

Our encoder $f$ is a six-layer temporal convolutional network (TCN) \cite{tcn} with residual and squeeze-and-excitation blocks \cite{resnet, squeeze-and-excite} detailed and visualized in Figure \ref{fig:encoder-architecture}.
Following SimCLR, our projection head is a multilayer perceptron (MLP) with one ReLU-activated hidden layer.

%% file: data.tex
\begin{table*}[t]
    \centering
    \caption{The six data sets used jointly to learn eye movement representations. \textit{Hz}: sampling frequency of the eye-tracker, e.g. 500 Hz = 500 time points recorded per second. \textit{Time (s)}: Time in seconds spent by the viewers looking at the stimuli.}
    \begin{tabular}{|c|p{3.5cm}|p{2cm}|p{2.25cm}|c|c|c|c|}
        \hline
         Data Set & Stimuli & Tasks & Eye-Tracker & Hz & Time (s) & Viewers & Samples \\
         \hline
         \hline
         \textbf{EMVIC} \cite{emvic} & normalized face images & free-viewing & Jazz-Novo & 1000 & 2.5 (ave) & 34 & 1430 \\
         \hline
         \textbf{FIFA} \cite{fifa} & indoor, outdoor scenes & free-viewing, search & SR Research EyeLink & 1000 & 2 & 8 & 3200 \\
         \hline
         \textbf{ETRA} \cite{etra, etra-2} & everyday scenes, puzzles & free-viewing, search & SR Research EyeLink II & 500 & 45 & 8 & 960 \\
         \hline
         \textbf{MIT-LR} \cite{mit-lowres} & outdoor scenes, pink noise & free-viewing & ETL 400 ISCAN & 240 & 3 & 64 & 12,352 \\
         \hline
         \textbf{MIT-LTP} \cite{mit-ltp} & outdoor scenes, portraits & free-viewing & ETL 400 ISCAN & 240 & 3 & 15 & 15,045 \\
         \hline
         \textbf{MIT-Search} \cite{mit-search} & outdoor scenes & search & ISCAN RK-464 & 240 & 1.2 (ave) & 14 & 12,768 \\
         \hline
         \hline
         Total & & & & & & \textbf{143} & \textbf{45,755} \\
         \hline
    \end{tabular}
    \label{tab:data-specifications}
\end{table*}
\section{Data}


We utilize six public data sets, having different specifications such as viewer demographics, experimental conditions, and equipment (listed in Table \ref{tab:data-specifications}). We stress that this further makes representation learning a non-trivial task due to their inherent variances. Nevertheless, we chose to experiment on this scale as this is necessary to evaluate the usability and generalizability of our method. Note that our work is limited to eye movements obtained from viewing static images. Eye movements obtained during reading, watching videos, or "in-the-wild" are out of scope.

To enforce some uniformity in our joint data set, we scale the coordinates such that the viewers' one degree of visual angle corresponds to 35 pixels (35px/dva). We work at a sampling frequency of 500 Hz. We downsample 1000Hz to 500Hz by dropping every other time point, and we upsample 240Hz to 500Hz by cubic interpolation.
We then opted to work with velocity signals as these were shown to be more meaningful \cite{gazemae}.
Note it is still possible to use our methodology on position signals if preferred.

%% file: experiment.tex
\section{Experiments}
\subsection{Training}
We train two networks: \textbf{CG}, which is trained on all available data and \textbf{CG-3}, a model trained only on EMVIC, FIFA, and ETRA data sets. We do this to have a fair comparison with the velocity models of GazeMAE (GM\textsubscript{v}) \cite{gazemae}, a previous work on eye movement representation learning using deep convolutional autoencoders.
CG and CG-3 employ all cropping methods and data transformations, and have the same network architecture described in Section \ref{sec:clrgaze} and Figure \ref{fig:encoder-architecture}.

Our inputs are 1-second segments or 500 time points ($T' = 500$). Our encoder $f$ then maps a sample signal $x \in \mathbbm{R}^{(2, 500)}$ to its representation $h \in \mathbbm{R}^{512}$.
We train our networks to minimize the NT-Xent Loss (Eq. \ref{eq:nt-xent}) with $\tau = 0.3$, learning rate=5e-4, batch size=1000 and Adam optimizer \cite{adam}. CG-3 is trained for 100 epochs while CG is trained for 800.
For a training set of 45,755 samples, this results to 45 batches per epoch. We train for 800 epochs or 36,000 iterations.
We notice that performance does not improve beyond this.
The networks are implemented with PyTorch 1.7 \cite{pytorch} and trained on an NVIDIA RTX 2080Ti GPU. We compute with automatic mixed precision (AMP) to enable larger batch sizes and faster training time.
All parameters are chosen empirically, and a random seed was set for all experiments.
Our code will be made available at https://github.com/chipbautista/clrgaze.

\subsection{Evaluation on Downstream Tasks}
We use the trained network to encode eye movements into feature vectors. We now input full-length samples $x \in \mathbbm{R}^{(2, T)}$ instead of fixed-length segments used for training. The GAP layer allows our encoder to handle arbitrary lengths, making it a more practical approach. The feature vectors are then used as inputs to a linear classifier, which is a standard evaluation method for representation learning \cite{rep-learning-bengio-2013}. In our case, our classifier is a linear Support Vector Machine (SVM) implemented through scikit-learn library \cite{scikit-learn}.

Limited by the available data labels, we opt to evaluate the learned representations by classifying the viewers based on their eye movements. This is also known as eye movement biometrics, a growing research area that, if done with traditional feature extraction methods, requires extensive knowledge on eye movements \cite{rigas-biometrics}.
When possible, we compare our results with other works that have conducted the same tasks.

\begin{table}[h!]
    \centering
    \caption{Accuracies achieved on biometric tasks using the learned representations as input to a linear SVM classifier.}
    \begin{tabular}{|p{2.5cm}|c|c|c|c|}
    \hline
      & \multirow{2}{*}{Others} & GM\textsubscript{v} & \textbf{CG-3} & \textbf{CG} \\
      & & \cite{gazemae} & (ours) & (ours) \\
     \hline
     EMVIC-Train & 86.0 \cite{lpitrack} & 86.8 & \textbf{94.2} & 92.7 \\
     \hline
     \multirow{3}{*}{EMVIC-Test} & 81.5 \cite{lpitrack} & \multirow{3}{*}{87.8} & \multirow{3}{*}{\textbf{94.5}} & \multirow{3}{*}{94.3} \\
     & 82.3$^{**}$ & & & \\
     & 86.4$^{**}$ & & & \\
     \hline
     EMVIC,ETRA, FIFA & - & 79.8 & \textbf{96.6} &  96.5 \\
     \hline
     ETRA & - & - & \textbf{96.0} & 95.0 \\
     \hline
     FIFA & - & - & 97.0 & \textbf{97.3} \\
     \hline
     MIT-LR & - & - & 60.6$^*$ & \textbf{82.9} \\
     \hline
     MIT-LTP & - & - & 74.0$^*$ & \textbf{90.5} \\
     \hline
     MIT-Search & - & - & 62.9$^*$ & \textbf{73.2} \\
     \hline
     All & - & - & 69.5$^*$ & \textbf{84.6} \\
     \hline
  \multicolumn{5}{l}{$^*$ data set not used in training}\\
   \multicolumn{5}{l}{$^{**}$ mentioned in \cite{lpitrack} but no citation was found}

    \end{tabular}
    \label{tab:main-results}
\end{table}

From Table \ref{tab:main-results}, it is shown that CLRGaze outperforms the previous works in all tasks.
We believe that this boost can be attributed to our methodology.
Recall that the contrastive CNN has to correctly classify if two segments originated from the same eye movement signal. To do so, it has to extract the patterns that are present throughout the signal, which are patterns that are likely to be idiosyncratic or unique to the viewer.
Applying random cropping and chunk transformations to the eye movements further encouraged the CNN to extract these global information patterns. This concept is related to slow feature analysis and contrastive predictive coding \cite{slow-features, cpc-infonce}.

Also, notice that substantially lower accuracies were achieved for the MIT data sets, which may indicate that 240Hz eye-trackers cannot capture granular information needed for biometric tasks.

%% file: results.tex
\subsection{Effect of cropping methods and data transformations}
Next, we train more models with the same parameters, changing only the composition of data transformations. We evaluate on the Biometrics (All) task since we deem this the most difficult. From Table \ref{tab:ablations-data-transformations}, it is shown that the choice of cropping methods and data transformations largely impacts accuracy. While we present only the results for Biometrics (All) for simplicity, note that we observed the same trend when evaluated on other tasks.

\begin{table}[h]
    \centering
    \caption{Accuracies achieved on the Biometrics (All) task by models trained with varying cropping methods and data transformations.}
    \begin{tabular}{|l|c|}
    \hline
     & Biometrics (All) \\
     \hline
     \multicolumn{2}{|c|}{\textit{Cropping Methods (refer to Figure \ref{fig:cropping-method})}} \\
     \hline
     Same & 75.3 \\
     Consecutive & 79.3 \\
     Random & 84.1 \\
     Consec, Same & 81.4 \\
     Random, Same & 84.9 \\
     Random, Consec & 83.2 \\
     \hline
     \multicolumn{2}{|c|}{\textit{Transformations (refer to Table \ref{tab:data-transformations})}} \\
     \hline
     None (\#1 only) & 78.9 \\
     Dropout (\#1-5) & 82.7 \\
     Dropout, Noise (\#1-7) & 83.9 \\
     \hline
     \hline
     Full model (\textbf{CG}) & 84.6 \\
     \hline
    \end{tabular}
    \label{tab:ablations-data-transformations}
\end{table}

\subsection{CLRGaze generalizes to unseen samples}
\begin{figure}[h]
    \centering
    \caption{t-SNE \cite{tsne} plots of the representations for validation samples, by data set. Point colors correspond to viewers.}
    \includegraphics[width=\columnwidth]{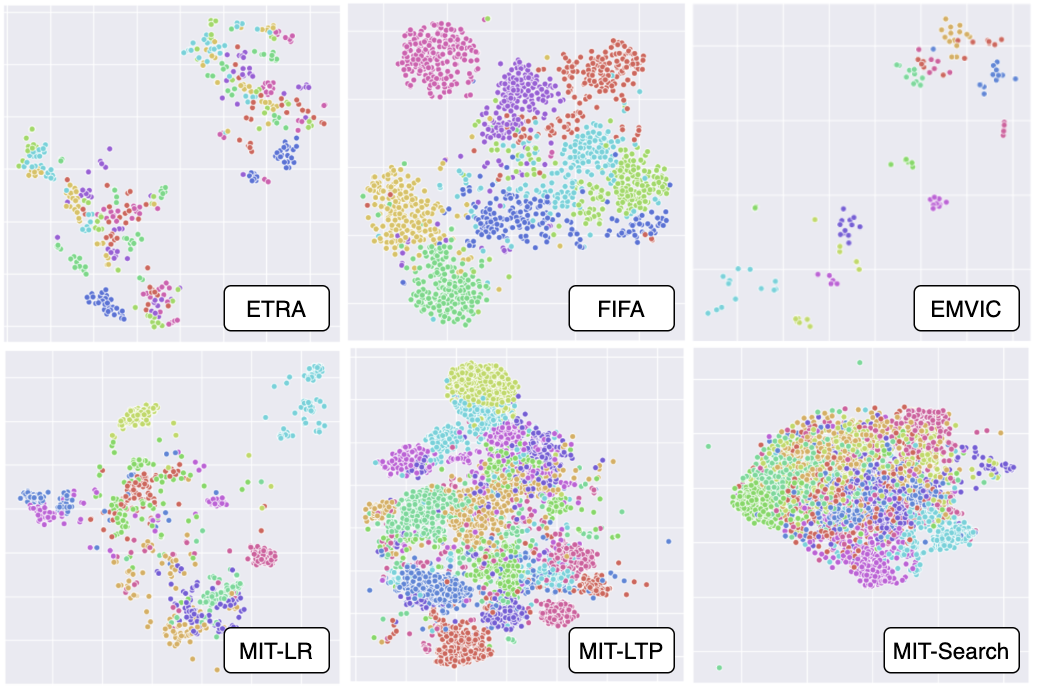}

    \label{fig:tsne_val}
\end{figure}

Finally, we train a model with the same parameters but on a viewer-stratified split (22,877 training and 22878 validation samples). In Figure \ref{fig:tsne_val}, we plot the representations of the validation set, using the viewers as the labels. Eye movements of a subject lie close together in the representation space, suggesting that our model can represent unseen samples sufficiently.

%% file: conclusion.tex
\section{Conclusion}
We take on a contrastive learning approach based on SimCLR \cite{simclr} to learn representations of eye movement signals. To port this methodology to the signal domain, we propose a set of data transformations that encourage a contrastive CNN to extract meaningful patterns from signals. We apply this methodology to six eye-tracking data sets despite varying specifications. The learned representations are evaluated with biometric tasks and a linear classifier, achieving high accuracies and outperforming previous works. Lastly, we show that the model can handle unseen samples well. This work presents a medium-scale experiment that advances eye movements-based deep learning applications.
